\documentclass{article}
\usepackage{ijcai17}
\usepackage{algorithmic}
\usepackage{algorithm}
\usepackage{graphicx}

\usepackage{tikz}
\usetikzlibrary{shapes}

\RequirePackage{ifthen}
\usepackage{latexsym}
\RequirePackage{amsmath}
\RequirePackage{amsthm}
\RequirePackage{amssymb}
\RequirePackage{xspace}
\RequirePackage{graphics}
\usepackage{graphicx}
\usepackage[small]{caption}
\usepackage{subcaption}
\usepackage{xcolor}
\usepackage{subcaption}

\RequirePackage{textcomp}
\usepackage{keyval}
\usepackage{xspace}
\usepackage{mathrsfs,paralist, amsmath,amssymb,url,listings,mathrsfs}
\usepackage{framed}
\usepackage{lipsum}
\definecolor{reddish}{rgb}{1,.8,0.8}
\definecolor{blueish}{rgb}{0.8,.8,1}
\definecolor{greenish}{rgb}{.8,1,0.8}
\definecolor{yellowish}{rgb}{1,1,.20}

\usepackage{times}

\title{CHARDA \\
Causal Hybrid Automata Recovery via Dynamic Analysis}
\author{Adam Summerville, Joseph Osborn, Michael Mateas\\
University of California, Santa Cruz\\
\{asummerv, jcosborn\} @ucsc.edu, michaelm@soe.ucsc.edu}

\begin{document}

\maketitle

\begin{abstract}
We propose and evaluate a new technique for learning hybrid automata automatically by observing the runtime behavior of a dynamical system.
Working from a sequence of continuous state values and predicates about the environment, CHARDA recovers the distinct dynamic modes, learns a model for each mode from a given set of templates, and postulates \textit{causal} guard conditions which trigger transitions between modes. Our main contribution is the use of information-theoretic measures (1)~as a cost function for data segmentation and model selection to penalize over-fitting and (2)~to determine the likely causes of each transition.
CHARDA is easily extended with different classes of model templates, fitting methods, or predicates. In our experiments on a complex videogame character, CHARDA successfully discovers a reasonable over-approximation of the character's true behaviors.
Our results also compare favorably against recent work in automatically learning probabilistic timed automata in an aircraft domain: CHARDA exactly learns the modes of these simpler automata.
\end{abstract}

\section{Introduction}

\textit{Hybrid automata} (HAs) combine discrete finite state machines with continuous variables~\cite{alur1993hybrid}. 
These continuous variables are updated at different rates in different \textit{states} (also called \textit{modes}) according to state-specific \textit{flow constraints}.
\textit{Transitions} between states may be \textit{guarded} on conditions involving (classically) the continuous variables or other predicates, and these transitions may \textit{update} continuous variables to new values instantaneously.
States may also have associated \textit{invariant} conditions; if an invariant is violated, the state  immediately exits along one of its available transitions.% (if no transitions are available, the automaton is in an error state).

Hybrid automata are a convenient notation for many different dynamical systems, and have at least semi-decision algorithms for a variety of interesting properties (e.g.\ satisfiability of LTL formulae, general reachability, and existence of optimal control policies)~\cite{alur1995algorithmic,henzinger1995decidable,henzinger1999discrete}.
%% MM: Citations for the above.
%% Joe: Done
Learning (or recovering) HAs from existing systems yields convenient abstractions for human analysis and high-level automated planning; moreover, these abstractions can be refined, possibly automatically (via new data or experimentation).

In this work we present \textit{CHARDA}, Causal Hybrid Automata Recovery via Dynamic Analysis, a non-parametric framework that learns an HA from  observations of a dynamical system.
CHARDA has two phases: mode identification and causal guard learning.
We identify modes via a dynamic programming approach that segments the trace and finds switchpoints where the dynamics of the system change.
Then CHARDA learns causal guard conditions for mode-to-mode transitions using information-theoretic measures.

CHARDA's segmentation requires no prior knowledge of the number of potential modes or the location of switchpoints, requiring only a set of potential model templates (e.g.\ $\dot{x} = a$ or $\dot{x} := a; \ddot{x} = b$, read respectively as constant velocity or constant acceleration $b$ starting from a reset velocity value $a$).
Although the models can take any form (so long as a likelihood function is available), here we use general linear models (multivariate linear regressions). 
CHARDA performs model selection and segmentation via a principled penalty function.
In this work, we tried both the Bayesian Information Criterion (BIC) and Minimum Description Length (MDL), but CHARDA is also penalty-function-agnostic. 

We demonstrate CHARDA in a novel domain: videogames, specifically \textit{Super Mario Bros} (SMB).
Games offer a unique set of challenges including non-physical dynamics and potentially very frequent mode transitions on the order of fractions of a second.
As a domain, games lie somewhere between synthetic data and a physical robot or other cyber-physical system.
%% MM: Since a video-game is a simulation, I'm confused as to how games "lie between fully simulated data and a physical robot". Need to say a bit more about this.
%% ADAM: Oops, that was supposed to say synthetic
Furthermore, games are interesting objects of analysis in their own right.
In games specifically, CHARDA has some exciting applications:
\begin{itemize}
\item In the General VideoGame (GVG) playing domain, an AI could derive HA models for game entities and then do planning on this abstracted space without relying on a forward model~\cite{perez2016general}
%% MM: GVG cite(s)
%% ADAM: I'll add
\item Model-checking/safety analysis of character automata without the overhead of manual modeling by human game designers~\cite{biped}
%% MM: AI-based game design cites which such check would be useful (e.g. Mark N. or Adam Smith cites, etc.). 
%% ADAM: I'll add
\item Extracting features for quantitative comparative analysis between games or game rules~\cite{fasterholdt2016you,ho2016roguelike}
%% MM: What's an example of the above? Is there existing work that is trying to do this? As an AI and games researcher, not sure what this is pointing at.
%% ADAM: I'm thinking the Jumping paper, Dylan's "What is Strafe Jumping?", there was a Roguelike one by Xavier Ho that's never been published, the combat paper/
\item Automatic scraping of characters from existing games for a character behavior corpus, which could then be used for analysis or procedural generation as game levels are already~\cite{summerville2016vglc}
%% MM: What would such automated scraping be used for? I assume the AI-based game design application? Need to say a bit more. 
%% ADAM: Expanded a touch
%% Joe: Added a cite for VGLC
\end{itemize}

The rest of the paper is structured as follows.
First, we discuss other approaches to learning dynamical system models and how CHARDA fits into the existing work here.
We then briefly introduce the concrete domain of interest and explain CHARDA's design and implementation.
Finally, we evaluate CHARDA in two domains: internally on the SMB domain, and externally in an aircraft tracking domain for comparison with another recent automaton learning algorithm.

\section{Related Work}

Hybrid automata are an attractive computational model for analysis, control synthesis, and estimation of real-world systems.
The inclusion of discrete behavior makes them expressive enough to describe many dynamical systems of interest, and although many classes of hybrid automaton have strong undecidability results~\cite{henzinger1995decidable} there are efficient semi-decision procedures to determine configuration reachability or equivalence between automata~\cite{alur1995algorithmic}.
Hybrid automata, suitably constrained, can also be directly implemented in software or hardware, with proofs about the model translating to the implemented system (given assumptions of e.g.\ component failure rates and latencies).

Despite the general undecidability of many HA properties, it is possible to constrain models or carefully choose semantics to obtain different analysis characteristics: discretizing time or variable values evades undecidability by approximating the true dynamics~\cite{jha2007symbolic}; keeping these continuous but constraining the allowed flow and guard conditions admits geometric analysis~\cite{frehse2005phaver}; and one can always merge states together to yield an over-approximation, producing smaller and simpler models.
There are also composable variations of hybrid automata that admit compositional analysis~\cite{alur2003hierarchical} as well as a logical axiomatization~\cite{platzer2008differential}, not to mention the body of tools and research that already exist for synthesizing control policies, ensuring safety, characterizing reachable areas, et cetera.

Given the desirable properties of this class of model, and the ready availability of tools for dealing with them, many researchers have explored automatically recovering these high-level models from real-world system behaviors.
CHARDA shares motivations with HyBUTLA~\cite{niggemann2012learning}, which also aimed to learn a complete automaton from observational data.
%% A superficial difference is that their work operates on multiple episodes of observations; CHARDA could easily be extended in that way.
%% MM: I'd delete the above sentence. Since it's a superficial distinction it's only a distraction. 
%% Joe: OK
HyBUTLA seems able to learn only acyclic hybrid automata, since it works by constructing a prefix acceptor tree of the modes for each observation episode and then merges compatible modes from the bottom up.
%% MM: added "for each observation episode" above since I'm suggesting deleting the previous sentence.  
%% Joe: OK
Moreover, HyBUTLA assumes that the segmentation is given in advance and that all transitions happen due to individual discrete events, presumably from a relatively small set.
The overall structure of both algorithms---split the observations into a number of intervals in which mode functions are fit, then merge redundant modes---is similar, but CHARDA learns a larger class of automata and does not require data to be pre-split into episodes or segments.

Santana \textit{et al.}~\shortcite{hybridmodels2015santana} learned Probabilistic Hybrid Automata (PHA) from observation using Expectation-Maximization.  At each stage of the EM algorithm a Support Vector Machine was trained to predict the probability of transitioning to a new mode. Unlike CHARDA, their work requires a priori knowledge about the number of modes.

The closest work to ours is that of Ly and Lipson\shortcite{ly2012learning} which used Evolutionary Computation to perform clustered symbolic regression to find common modes with the Akaike Information Criterion uses to penalize model complexity.  However, unlike CHARDA their work assumes \textit{a priori} knowledge about the number of modes. Moreover, since their work assigns individual datapoints, not intervals, to a mode, their approach can only model stationary processes.

Several approaches have sought to learn models that describe dynamical systems' behavior.
Hidden Markov Models~\cite{baum1966statistical} learn probabilistic state transitions between a hidden state and the observed data.
The Infinite HMM ~\cite{beal2002infinite} extends this to an unbounded number of states which assumes a Chinese Restaurant Process governs the state space.
These approaches do not characterize guard \textit{conditions}, but instead learn the \textit{probability} of taking state transitions at each instant.

Data segmentation has a natural connection to automaton learning, and CHARDA uses an approach based on least squares regression~\cite{bellman1969curve}.
Model-based recursive partitioning~\cite{zeileis2008model} is an alternative family of techniques which fits a model to the entire dataset and then iteratively and greedily splits that model until reaching a threshold quality level or split count.
Unfortunately, each split is only locally optimal so there are no guarantees about global optimality.
The Forget-Me-Not-Process~\cite{milan2016forget} finds a partitioning of time segments that allows for models to be repeated across different partitioned segments; however, it only works for stationary processes, i.e.\ distributions that do not change over time.

% Some techniques have leveraged certain inductive biases from biological or other domains to obtain good results.
% Kukreja \textit{et al.}~\shortcite{kukreja2005least} found models of switched mode systems given prior knowledge about the locations of the switchpoints, and Bridewell \textit{et al.}~\shortcite{bridewell2008inductive} used exhaustive search over model structures to best explain observed data without assuming fixed switchpoints.
% CHARDA takes advantage of inductive biases around the behaviors of embodied agents, learning not just what dynamical modes there are but also \textit{why} the character switches from one to another.

In terms of finding abstract models specifically of Nintendo games, we were inspired by Murphy's work~\shortcite{murphy2016glend} in automatically determining physical properties of game characters.
That project, like ours, examined runtime memory structures to determine where objects were; they further explored, through experimentation, causal linkages between arbitrary locations in RAM and the visual position of characters on the screen.
These relations were used to drive other experiments, e.g.\ to discover whether game characters fell due to gravity or whether their movement was obstructed by particular types of game objects. 
In a sense, their work is an ad hoc
%% MM: assuming *this* refers to Murphy's work, not ours, moved the sentence before the one about our work as the referent for *this* was confusing otherwise. 
%% Joe: Reworded a bit to make it clearer.
property-based testing approach to learning which of a fixed set of properties holds.
Our work requires less domain knowledge and captures the characters'  behavior more precisely. 

In the future we look forward to combining our more general approach with such knowledge-rich techniques to capture more complicated interactions between multiple agents and their environment.
A recent publication by Summerville \textit{et al.}~\shortcite{qmark2017summerville} similarly used games as their domain, attempting to find causal interactions shared by different entities, and we build on this approach for the causal guard learning. 

%HAs can alternately be viewed as the combination of a conventional finite automaton, a set of switched differential equations contingent on the current state, and an event generator which sends input tokens to the finite automaton.

\section{Domain}
CHARDA learns hybrid discrete/continuous behaviors of videogame characters or other agents whose inputs and movement behavior are observable.
We obtain these inputs from an example playthrough of a game (e.g. SMB), assuming these inputs are representative of the character in question.
Replaying this input sequence once through a software emulator of the game's hardware platform, we read out high-level features from the simulated graphics hardware and assemble those into distinct agents whose positions are tracked over time (we elide the details for space).
Importantly, characters may pop in and out of existence, collide with fixed or moving obstacles of various types, or perform other arbitrary (often non-physical) behaviors.
We can only observe characters' positions at a resolution of 1 pixel (a character is generally 8--32 pixels high); even then, the game world and our sensing are at a \(\frac{1}{60}\)-second fixed discrete time interval.
All our position readings are therefore inaccurate by up to one spatial unit, and these errors naturally propagate to velocity and other calculations.

The input to our automaton learning process %(obtained by the above process) 
for a single entity is: sequences of discrete variable values that are possible control inputs (e.g.\ button presses), continuous variable values,
%I removed this since it's only low precision for Mario, we can expand on that later if we want
%at low precision, 
and sets of predicates describing facts in external theories such as
%% MM: why "at low precision"? Where is the low-precision aspect coming from?
%% Joe: Yes, you're right
collision (e.g., the character was touching an object with appearance \(A\) at time \(t\) on one side or another).
The goal is to go from that input data, presumed to be representative of the entity's ``true'' behaviors, to an abstraction suitable for planning or other purposes.
This type of data is not hard to obtain for cyber-physical systems under the analyst's control or in cases where the possible causes for behavior change can be observed at some precision (even a probability distribution for these causes would suffice).

In this work, we look at learning a constrained class of hybrid automata from a combination of controlled (or at least witnessed) inputs and observed outputs.
Specifically, though the learned automata may have any structure in terms of the number of modes and transitions, the modes may only have flows from a given set of model \textit{templates}.
In this specific work (and without loss of generality), every mode's flow condition is a specialization of \(\ddot{x} = a, a \in \mathbb{R}\)
%either \(\dot{x} = c\) or \(\ddot{x} = a\), for some real parameters \(c\) and \(a\)
; moreover, all transitions leading into a given mode are forced to have the same update function, either \(\dot{x} := n, n \in \mathbb{R}\) %for some real parameter \(n\) 
or the empty update.
%% MM: I assume for modes governed by \(\dot{x} = c\) you'd never have an edge update of \(\dot{x} := n\) where c <> n (that is, setting velocity to some value and then instantaneously setting it to a different value once you're in the mode)? The way this is written it sounds like such a situation is allowed in your space of models, but I can't see when you'd ever want that. 
%% Adam - Addressed
%% Joe: Yes, perfect fix
Finally, the set of guard conditions is currently assumed to be conjunctions of predicates from a given labeled set.
Our causal learning component learns which of these predicates is most associated with the transitions, and prefers those predicates which are more strongly causal.
%% MM: "labeled as more strongly causal": Even after reading the paper, I'm not sure if "labeled" here refers to the preference of exogenous predicates to endogenous predicates in edge conditions or to the computation of npmi. Perhaps just get rid of "labeled" here and say "predicates which are more strongly causal." 
%% Adam - did that
There is no reason these guards could not also be learned as e.g.\ linear inequalities, since we know the set of modes and their active intervals at the time of cause assignment.
Again, we focus here on learning reasonably small over-approximations of the true model: these can always be refined, but we don't want to exclude any witnessed behaviors.

\section{Method}

We break down the hybrid automaton learning process into two parts: Identifying modes and determining causes for transitions.
Again, these algorithms operate over a sequence of continuous variable values and a sequence of sets of predicates describing the automaton's environment at each instant.  We roughly follow the classic dynamic programming solution to the segmented least squares problem~\cite{bellman1969curve} with a number of distinctions:

\begin{itemize}
\item Different model templates are considered for each segment, instead of a single least squares regression
\item A principled penalty  instead of a hand-chosen constant
\item Merging segments if it results in a more optimal model.
\end{itemize}

\subsection{Mode Identification}
%% MM: I'm not following this whole section very well. What background are you assuming the reader has? As a "general AI person" but not a specialist in segmentation, I'm following the general outlines of this but not the mathematical details. Potentially you will have issues with IJCAI reviewers, though, ironically, with some reviewers, they fact they're not following the details will be positive because they won't want to admit they're not following and the fact there's lots of equations will lead them to say "this must be science". But I'm not such a reviewer.. ;)
% Adam - Hmm, I'm not sure, Joe?
The mode identification process first requires the construction of all possible models for all possible sub-intervals.  Let \(T\) be a table of model parameters with one entry for each interval \(i,j\) and model template \(m\).  Then we define \(T\)'s entries as:
\begin{gather*}T[i,j,m] = \mathrm{train}(m, d[i:j]) \quad \text{s.t.}\\
\quad 1 < i < j < n, m \in \mathcal{M}
\end{gather*}

where $n$ is the number of potential switchpoints, $\mathcal{M}$ is the set of model templates,  $d$ is the dataset, and $T[i,j,m]$ is the model of template $m$ trained on data from the interval of $i$ to $j$.  For this work our set of models are all multivariate regressions, but our approach is general enough to work with any approach that supports a likelihood function $\mathcal{L}(m|d)$.

The cost for a given model $m$ for sub-interval $i$ to $j$ is therefore:
\begin{gather*}C[i,j,m] = -\log (\mathcal{L}(T[i,j,m]|d[i:j])) + \mathrm{pen}(m,d[i:j])
\end{gather*}

given the penalty criterion \(\mathrm{pen}\). For this work we considered two penalties for model complexity.  We wanted a principled measure for model complexity for the selection of a given sub-model for an interval, for when a break should occur (due to the inclusion of a switch point increasing model complexity), and for when a merging of modes should occur (due to the inherent fact that two similar but distinct modes are more complex than one mode).  To that end we considered both the Bayesian Information Criterion (BIC)~\cite{schwarz1978estimating} and the Minimum Description Length (MDL) ~\cite{stine2004model}. \vspace{-0.25\baselineskip}
\begin{gather*}
\mathrm{BIC} = -\log(\mathcal{L}(m|d)) +\mathrm{dim}(m)\log(n)/2\\
\text{So: }
\mathrm{pen}_\mathrm{BIC} =\mathrm{dim}(m)\log(n)/2
\end{gather*}
Where $\mathrm{dim}(m)$ is the number of parameters in model $m$ and $n$ is the number of datapoints in dataset $d$. \vspace{-0.25\baselineskip}
\begin{gather*}
\mathrm{MDL} = -\log(\mathcal{L}(m|d)) +\mathrm{dim}(m)(1 + {\log(n)}{/2})\\
\text{So: }
\mathrm{pen}_\mathrm{MDL} = \mathrm{dim}(m)(1 + {\log(n)}{/2})
\end{gather*}
The two measures are very similar, being asymptotically the same, but differ in the constants applied to the penalty term.  BIC assumes a Bayesian standpoint and determines which model from a set of models is the true model.  It operates asymptotically as $n$ trends to $\infty$, given a fixed loss for choosing the wrong model.  MDL instead takes an information theoretic standpoint and assumes a spike-and-slab prior distribution for each parameter.  Given that prior it takes approximately $(1 + \frac{\log(n)}{2})$ bits to encode the parameter, 1 bit for whether the parameter $\theta = 0$ (i.e.\ is a slab) and $\frac{\log(n)}{2}$ bits to encode its value (i.e.\ if it is a spike).

For all segments that end at point $j$ we find the optimal model and segmentation that leads to that point. $O[j]$ is the optimal cumulative cost of models across segments up to datapoint $j$.
\begin{gather*}
O[j] = \left\{
\begin{array}{ll}
      0 & \text{if }j = 0 \\
      \mathrm{argmin}_{0 \leq i \leq j,m} \,\,\, C(i,j,m) + O[i-1] & \text{if }j > 0 \\
\end{array} 
\right.
\end{gather*}

We use dynamic programming to work backwards from the last switch point, finding the optimal sequence of segments that produces the optimal set of models,

After segmentation, the segments' models are merged if this will improve the overall attractiveness of the entire model, namely by reducing the number of parameters in the overall model by a large enough amount that the decrease in complexity is greater than the decrease in likelihood. 

This is accomplished by constructing a new model from the data for segments $s_1, s_2$ concatenated to $d[s_1s_2]$:
\begin{gather*}
m_{s_1s_2} = \mathrm{argmin}_{m \in \mathcal{M}} \,\, \log \mathcal{L}(\mathrm{train}(m, d[s_1s_2]) | d[s_1s_2])
\end{gather*}
The overall sequence of models is improved by the merging if the following inequality holds:
\begin{gather*}
-\log(\mathcal{L}(m_{s_1s_2}|d[s_1s_2])) + \mathrm{pen}(m_{s_1s_2}) < \\-\log(\mathcal{L}(O[s_1])) - \log(\mathcal{L}(O[s_2])) + \\ \mathrm{pen}(O[s_1])+ \mathrm{pen}(O[s_2])
\end{gather*}

\subsection{Guard Learning}

From these merged modes, causal guarded transitions between modes are learned by finding probabilistically likely conditions where the direction of causality is known.
Our target domain comes with some advantages for ascribing causality, namely we have inputs supplied by a player and we can be sure of the direction of causality regarding them; however, any domain that allows for instrumentation of exogenous inputs can utilize our same methodology.
Another potential source of causal transition guards in our domain is collisions between visible entities, of which, again, we can be sure of the direction of causality.
We also look at endogenous variables as a last resort (and then mainly qualitatively), since causality is much harder to ascertain: for example, if we enter a mode with flow $\dot{y} = -4$ it could be that $\dot{y}$ is saturating at a terminal velocity, or it might be for some other reason.

For the SMB domain we consider the following set of predicates for guard condition learning:
\begin{itemize}
\item \textit{Control} $I$ (Pressed; Held; Released) --- A change in the binary control input $I$ --- \textit{Exogenous}
\item \textit{Collision} with $X$ from direction $Y$ --- Collision with another entity, $X$, from a given direction $Y$ --- \textit{Exogenous}
\item \textit{$0$-in, $0$-out by Sign} - A zero crossing or touching in velocity and its characteristics (e.g.\ from negative to positive, or vice versa) --- \textit{Endogenous}
\item \textit{Velocity Extremum} - $\dot{x} = \mathrm{ext_{\pm}}(s)$ - the velocity is roughly equal to the extremum for a given mode $s$  --- \textit{Endogenous}
\item \textit{Acceleration Sign} --- $\ddot{x}$ has the sign -1, 0, or 1  --- Endogenous
\item \textit{Velocity Sign} --- $\dot{x}$ has the sign -1, 0, or 1 --- \textit{Endogenous}
\end{itemize}
The \textit{Control} and \textit{Collision} predicates are given priority as we can be sure of their direction of causality.

Summerville \textit{et al.}\ used Normalized Pointwise Mutual Information (NPMI) to learn semantic information about game objects~\shortcite{qmark2017summerville}, which led us to believe that we could determine transition guards using a similar technique.
We calculate the NPMI of each transition from a predecessor mode to a successor mode with each predicate active during the  predecessor mode.
NPMI is a scaling of pointwise mutual information defined as:
\begin{gather*}
\mathrm{npmi}(x,y) = \frac{\mathrm{pmi}(x,y)}{- \log( p(x,y))} = \frac{\log (\frac{p(x,y)}{p(x)p(y)})}{- \log( p(x,y))}
\end{gather*}
NPMI for two events is $-1$ when they never co-occur, $0$ when  independent, and $1$ when they always co-occur.
In this work we considered two different thresholds for NPMI, $0.9$ for \textit{universal} (present all, or nearly all, the times that transition is taken) events and $0.4$ for \textit{relevant} events.
For example, to learn the cause for transitions from hypothetical mode \textbf{A} into mode \textbf{B}, we look at all time intervals where \textbf{A} is active, determine for each predicate how strongly correlated it is with the transition event \(\mathbf{A}\Rightarrow\mathbf{B}\), and take all those passing a threshold to be causes.
These correspond to conjuncts in the guard condition. Those correspondences which are high enough to be of interest but do not meet the threshold are called \textit{relevant} and are possible disjuncts in the guard condition (assuming it has the form \(c_1 \land \ldots \land c_i \land (d_1 \lor \ldots \lor d_j)\)).
If we have an exogenous explanation, we discard endogenous explanations.
%% MM: What does "sufficiently" mean to discard the endogenous explanations? And just to be clear, the exogenous predicates are Control and Collision, correct? 
%%Adam: Addressed by both labeling exogenous/endogenous and fixed up sufficiently -> sufficient

We may have cases where out-transitions of a mode are non-deterministic: they have identical causes, or one's causes subsume another.  
In these situations the offending target modes are merged, one pair at a time, re-connecting edges as necessary until a fixpoint is reached.
%% MM: I assume "any such modes" means the target modes (the two different modes being transitioned into via non-deterministic out-transitions) not the source mode? Say a little something to disambiguate this. 
%% Joe: Tweaked a little and added another sentence
This merging greedily abstracts the true automaton, but in practice it seems to work well for domains like game characters whose discrete state changes are generally strongly tied to control inputs or collisions; future work will explore more sophisticated approaches to resolving non-determinism.

\section{Evaluation}

To evaluate our work we considered two domains: Aircraft Dynamics Modeling and Mario's Jump Dynamics from SMB.

We explore the use of CHARDA in aircraft modeling for a direct comparison with Santana \textit{et al.}~\shortcite{hybridmodels2015santana}.
Their approach used Expectation Maximization~\cite{dempster1977maximum} to recover a hybrid automaton from observational data by iteratively refining an Interactive Multiple Model.
Guard conditions were learned by applying support vector machines. As in Santana, we also include results for a Jump Markov Linear System (JMLS) which assumes Markovian transitions.
%% MM: This last sentence feels like a disjunction. The whole paragraph is a meaty discussion of distinctions with HyBUTLA re: mode identification, then in the last sentence you switch away from both HyBUTLA and mode identification to say a non-sequitur about guard condition learning. It's also not related to your approach to learning guard conditions in any way. 
%% Joe: I moved it over here!

\begin{figure}[bt]
\centering
\includegraphics[width=0.8\columnwidth]{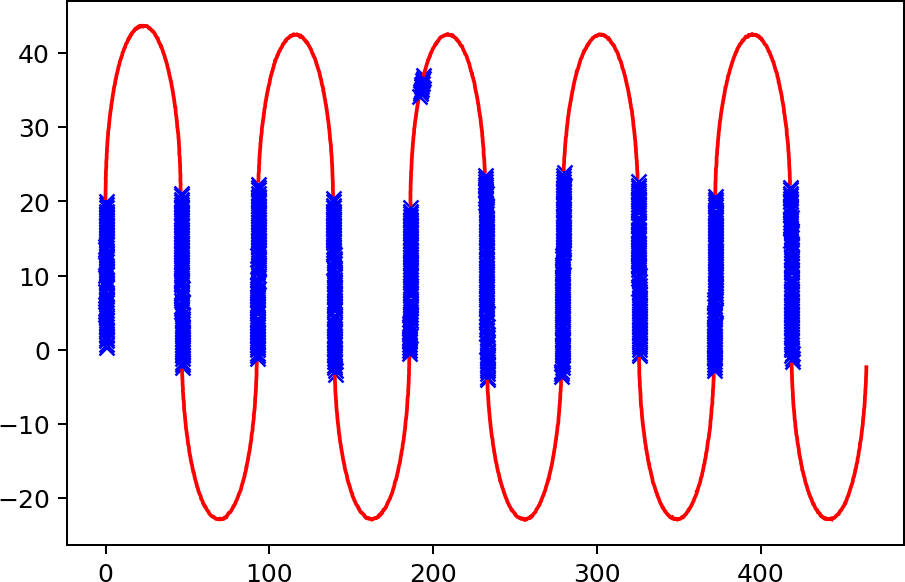}
\caption{\small  The Lawnmower data, as segmented by CHARDA with BIC.  Beyond having slight errors on the beginning and end of the turns, there is one turn where it incorrectly reverts to a constant velocity in the middle. *Only the switch point detection portion of CHARDA is used.}
\label{fig:lawnmower}
\end{figure}

The aircraft model is given in two distinct scenarios: the first, ``Lawnmower'' (see Fig.~\ref{fig:lawnmower}), features an aircraft moving in a constant velocity for some period of time and then making a constant-rate turn to reverse heading, repeating this pattern for some number of iterations.
In the second scenario, ``Random,'' the aircraft makes a given maneuver (either constant heading or constant turn) for 50 time steps and then changes to a random maneuver; this is repeated 17 times.
We must note that this portion of our evaluation is only based on CHARDA's segmentation algorithm and does not employ transition guard learning.
As the observational data offers no causal information indicating why a mode transition might be made, we do not learn any causal transition guards (which would simply overfit the given observations). 

As in Santana's work we ran 32 trials and discarded the best and worst runs; the results are shown in Table \ref{tab:results}.  We see that for the Lawnmower domain that we outperform Santana \textit{et al.}, but both are close enough to the ground truth that the difference is negligible.  In the Random domain we outperform the prior work dramatically because our segmentation is not based on learning linear guards; we instead find an optimal segmentation based on model accuracy and complexity.  We must note again that there are no real causes for why the aircraft changes maneuvers, so it is impossible to learn true causal guards.  Santana \textit{et al.}\ learn correlative guards for a given training instance, but their learned guards are not applicable to unseen data because they are tuned to that specific training instance (for example, if the aircraft's flight pattern was rotated or translated, all of their learned guards would be invalidated due to their training domain and linear nature).  As such, we feel that it is only relevant to compare the segmentation portion of CHARDA to the prior work. CHARDA would be better-suited if the domain were framed as a control problem and the dataset contained features like operator controls and aircraft sensors.

% However, given that no possible guards can be learned, the HA learned by CHARDA (or any other HA learning approach) would not be applicable to future unseen data, so this is not the most useful test.
%% MM: I'm a little confused. In the previous paragraph you say that "there are no causal reasons for why a transition is made" so there is no causal learning. I took this to mean that in this domain there are no causes. But now it sounds like PHA *does* learn guards, so there are predicates available for learning guards (I assume these are endogenous conditions like the last four in your Mario list). So then why didn't CHARDA use the same ones for learning conditions? Imagine a reviewer was thinking the following after reading this: "Hmmm, they're claiming they do better on the Random aircraft movement, but because they didn't learn conditions, they just let their model arbitrarily overfit to this exact trajectory, where PHA has learned a generalizable automaton. So CHARDA doesn't win at all." What could you say here to disarm such a reviewer (correcting a misunderstanding in this interpretation of the results, etc.). 

%%Adam:  So, the answer is - because they overfit too - they didn't have any test/training split and only reported training error - Their "learned" transitions were only valid for their given trajectory as well 

\begin{table}
\centering
\begin{tabular}{ l | c |r }
Method & Data & Attribution Error \\
\hline 
JMLS & Lawnmower & 53.93\%\\
  PHA & Lawnmower & 3.33\%\\
  CHARDA & Lawnmower & \textbf{2.75\%$^*$} \\
  \hline
JMLS & Random & 58.91\%\\
  PHA & Random &63.2\% \\
  CHARDA & Random & \textbf{4.10\%$^*$} \\
\end{tabular}
\caption{\small Percentage of modes misattributed for CHARDA, PHA, and JMLS.  The results shown here are only based on the segmentation portion, and do not include causal guard learning as there are no causal reasons for the mode transitions.}
\label{tab:results}
\end{table}

\begin{figure}[htb!]
\fbox{\parbox{\columnwidth}{
% \begin{center}
% \textbf{Reverse Engineered HA}
% \end{center}
\begin{description}
\item [On Ground] $\dot{y} = 0$ --- Caused by Mario colliding with something solid from above 
\item [Jump(1,2,3)] Three jumps with parameters: \begin{itemize}
\item $\dot{y} := 4, \ddot{y} = -\frac{1}{8} $
\item $\dot{y} := 4, \ddot{y} = -\frac{31}{256} $
\item $\dot{y} := 5, \ddot{y} = -\frac{5}{32} $ 
\end{itemize}  Entered from \textbf{On Ground} when the \textbf{A button} is pressed and $|\dot{x}| < 1$,  $1 \leq |\dot{x}| < 2.5$, or  $2.5 < |\dot{x}| $, respectively
\item [Release(1,2,3)]  $\dot{y} := min(\dot{y}, 3)$ --- Entered from the respective \textbf{Jump} when the \textbf{A} button is released; \(\ddot{y}\) same as respective \textbf{Jump}.
\item [Fall(1,2,3)]  Falling at one of three rates: $\ddot{y} = -\frac{7}{16}$, $-\frac{3}{8}$, or $-\frac{9}{16}$; entered from the respective \textbf{Jump} or \textbf{Release} mode when the apex is reached (\(\dot{y} \leq 0\))
\item [Terminal Velocity(1,2,3)] $\dot{y} = -4$ - Entered from \textbf{Fall} when $\dot{y} \leq -4$.  The initial timestep in the \textbf{Terminal Velocity} state is actually $\dot{y} = -4 + \dot{y}-\lfloor \dot{y} \rfloor$ before being set to $-4$.
\item [Bump(1,2,3)] $\dot{y} := 0$ --- Entered from a \textbf{Jump} or \textbf{Release} when Mario collides with something hard and solid from below; \(\ddot{y}\) same as respective \textbf{Jump} or \textbf{Release}
\item [SoftBump(1,2,3)] $\dot{y} := 1 + \dot{y}-\lfloor \dot{y} \rfloor$ --- Entered from a \textbf{Jump} or \textbf{Release} when Mario collides with something soft and solid from below; \(\ddot{y}\) same as respective \textbf{Jump} or \textbf{Release}
\item [Bounce(1,2,3)] $\dot{y} := 4, \ddot{y} := a$ --- Entered when Mario collides with an enemy from above; \(a\) is given by the respective \textbf{Jump}, \textbf{Release}, \textbf{Fall}, or \textbf{Terminal Velocity} state
\end{description}
}
}
\caption{\small The true HA for Mario's jump in \textit{Super Mario Bros.}  $:=$ represents the setting of a value on transition into the given mode, while $=$ represents a flow rate while within that mode. %Note that our given constrained learnable class is not expressive enough to learn it precisely.
}
\label{fig:TrueHA}
\end{figure}

For the Mario domain, we made no assumptions about the number of true modes and let the non-parametric nature of our approach attempt to recover the correct modes.  This means that we are unable to compare to Santana \textit{et al.}\ as it requires the number of modes a priori, so instead we compare our results to a manually-defined automaton based on human reverse-engineering of the game's program code~\cite{smbphysics} (see Fig.~\ref{fig:TrueHA}).
%% MM: Something wrong with citation. More substantively, what does "reverse engineered" mean? How do you get your hands on (compute) the "ground truth" automaton?
%% Joe: Mentioned the human element.
We present the HAs learned by CHARDA in Figure \ref{fig:LearnedHA}.  The Mario trace used for this work was 3772 frames in length, $~63$ seconds.  The learned HAs are over-approximations of the true HA.  Whereas the true HA has 3 separate jump modes based on the state of $\dot{x}$ at the time of transition, the learned HAs have only one such jump whose parameters are averages of the parameters of the true modes.  Following from learning just one jump, CHARDA learns only a single falling mode.  MDL does learn that releasing the \textbf{A} button while ascending leads to a different set of dynamics, but it considers this a change in gravity as opposed to a reset in velocity.

\begin{figure}
\centering
\includegraphics[width=0.95\columnwidth]{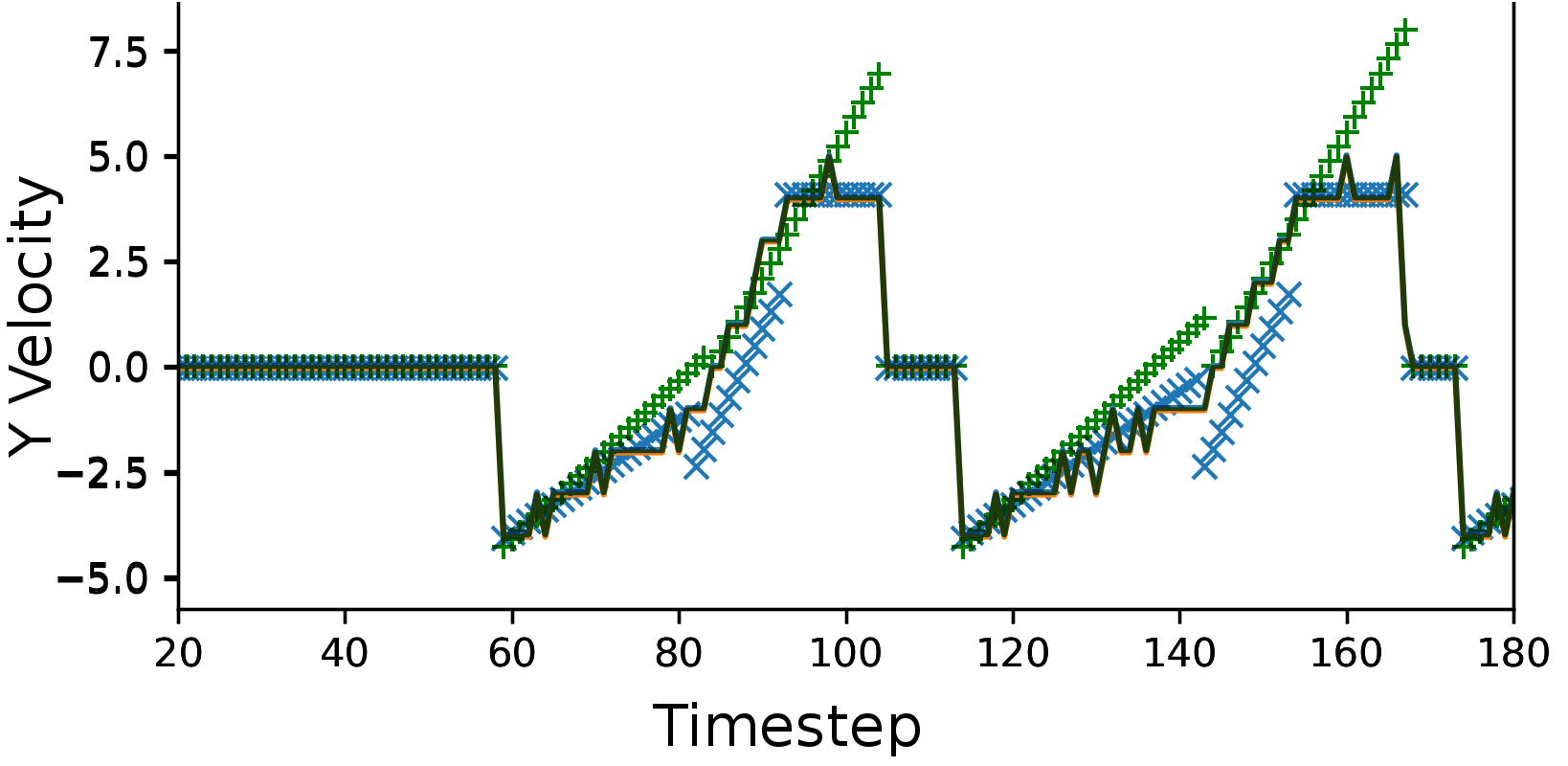}
\caption{\small Modeled behavior using MDL criterion (Blue X) and BIC (Green $+$) vs true behavior (Black Line). MDL's  largest error source is resetting to an specific value when the true behavior involves clamping to that value, whereas since BIC learns to transition at the 0 crossing it has a more accurate reset velocity.  BIC does not learn the transition from \textbf{Falling} to \textbf{Terminal Velocity}.  MDL has a Mean Absolute Error (MAE) of 0.522 while BIC has an MAE of 0.716.  }
\label{fig:mdl_vel}
\end{figure}

MDL produces the more faithful model of the true behavior, but is overzealous in its merging of the distinct jump mode chains into a single jump mode chain. As such, it only recovers 7 of the 22 modes; however, abstracting away the differences between the jump chains it learns 7 of 8 modes, only missing the distinction between hard bump and soft bump.  A comparison of the modeled behaviors and the truth can be seen in figure \ref{fig:mdl_vel}.

\begin{figure}[htb!]
\begin{subfigure}[b]{0.5\textwidth}
\fbox{\parbox{\columnwidth}{
% \begin{center}
% \textbf{MDL Criterion Recovered HA}
% \end{center}
\begin{description}
\item [On Ground] $\dot{y} = 0$ --- Caused by Mario colliding with something solid from above 
\item [Jump] $\dot{y} := [3.97,4.10], \ddot{y} = [-0.140,-0.131] $  --- Entered from \textbf{On Ground} when the \textbf{A button} is pressed
\item [Release] $\dot{y} := [2.10,2.54], \ddot{y} = [-0.430,-0.384] $  --- Entered from \textbf{Jump} when the \textbf{A button} is released
\item [Fall] $\dot{y} := 0, \ddot{y} = [-0.373,-0.359] $  --- Entered from \textbf{Jump} or \textbf{Release }when the apex is reached
\item [Bump]  $\dot{y} := [-1.85,-1.27], \ddot{y} = [-0.324,-0.238]$ --- Entered from \textbf{Jump}  when something solid is collided with from below
\item [Bounce]  $\dot{y} := [3.51,3.82], \ddot{y} = [-0.410,-0.378]$ --- Entered from \textbf{Jump}  when an enemy is collided with from above
\item [Terminal Velocity] $\dot{y} = [-4.15,-4.06]$ --- Entered from \textbf{Jump} or \textbf{Fall} 
% \item [On Ground] $\dot{y} := 0, \ddot{y} := 0$ --- Caused by Mario colliding with something solid from above 
% \item [Jump] $\dot{y} := 4.21 , \ddot{y} := -0.168 $  --- Entered from \textbf{On Ground} when the \textbf{A button} is pressed
% \item [Release] $\dot{y} := 0.617, \ddot{y} := -0.294 $  --- Entered from \textbf{On Ground} when the \textbf{A button}
% \item [Fall]  $\dot{y} := -0.484, \ddot{y} = -0.274$ --- Entered from \textbf{Jump} when the apex is reached or when collision with something solid from above is ended
% \item [Bump]  $\dot{y} := 0, \ddot{y} = -0.363$ --- Entered from \textbf{Jump}  when something solid is collided with from below
% \item [Bounce]  $\dot{y} := 4.00, \ddot{y} = -0.369$ --- Entered from \textbf{Jump}  when an enemy is collided with from above
% \item [Terminal Velocity] $\dot{y} := -4.105$ --- Entered from \textbf{Jump} or \textbf{Fall} 
\end{description}
}
}
\label{fig:MDL_HA}
\caption{\small HA with MDL as the penalty.}
\end{subfigure}
\begin{subfigure}[b]{0.5\textwidth}
\fbox{\parbox{\columnwidth}{
% \begin{center}
% \textbf{BIC Criterion Recovered HA}
% \end{center}
\begin{description}
\item [On Ground] $\dot{y} = 0$ --- Caused by Mario colliding with something solid from above 
\item [Jump] $\dot{y} := [4.19,4.42], \ddot{y} = [-0.195,-0.181] $  --- Entered from \textbf{On Ground} when the \textbf{A button}  is pressed
\item [Fall]  $\dot{y} := 0, \ddot{y} = [-0.356,-0.338]$ --- Entered from \textbf{Jump} when the apex is reached
\item [Bump]  $\dot{y} := [-2.37,-1.67], \ddot{y} = [-0.289,-0.188]$ --- Entered from \textbf{Jump}  when something solid is collided with from below
\item [Bounce]  $\dot{y} := [3.52,3.88], \ddot{y} = [-0.424,-0.391]$ --- Entered from \textbf{Jump}  when an enemy is collided with from above
\item [Terminal Velocity] $\dot{y} = [-4.16,-4.05]$ --- Entered from \textbf{Jump} when the threshold of $-4$ is reached.
% \item [On Ground] $\dot{y} := 0, \ddot{y} := 0$ --- Caused by Mario colliding with something solid from above 
% \item [Jump] $\dot{y} := 4.37, \ddot{y} := -0.202 $  --- Entered from \textbf{On Ground} when the \textbf{A button}  is pressed
% \item [Fall]  $\dot{y} := -0.484, \ddot{y} = -0.274$ --- Entered from \textbf{Jump} when the apex is reached or when collision with something solid from above is ended
% \item [Bounce]  $\dot{y} := 3.37, \ddot{y} = -0.410$ --- Entered from \textbf{Jump}  when an enemy is collided with from above
% \item [Terminal Velocity] $\dot{y} := -4.05$ --- Entered from \textbf{Jump} or \textbf{Fall} 
\end{description}
}
}
\label{fig:BIC_HA}
\caption{\small HA with BIC used as the penalty}
\end{subfigure}
%\caption{Learned Mario HAs. Parameters as 95\% CI.}
\caption{\small Learned Mario HAs. Parameters as 95\% confidence intervals.}
\label{fig:LearnedHA}
\end{figure}

\section{Conclusion and Future Work}

We have presented CHARDA, a novel combination of techniques (dynamic programming with a grounded penalty for data segmentation, causal relationship learning) that can recover hybrid automata from observations of a dynamical system.  CHARDA outperforms an existing HA learning algorithm in data segmentation, and in a well-suited domain can find causal (not merely correlative) transition guards.  We have also demonstrated CHARDA in a novel domain, videogames, that comes with an interesting set of challenges (short time durations, non-physical dynamics) and benefits (full access to all command inputs).

The use of a well-founded penalty criterion in conjunction with the dynamic programming approach is only one of many possible segmentation techniques, and it remains future work to test the general framework of \textit{Segmentation} + \textit{Guarded Transition} learning with other techniques.  However, the biggest source of error in the learned HAs comes not from mistakes in segmentation, but rather from overzealous merging of modes.  The learned parameters at segmentation in fact do describe modes in line with \textbf{Jump1} and \textbf{Jump3} (i.e.\ $\dot{y} = 4$ vs $\dot{y} = 5$), but these modes are merged together since it improves the overall learned model according to the criterion.  It remains for future work to determine if there is a different principled way to learn these similar but distinct modes.  It is also future work to incorporate techniques from other approaches, such as mode assignment via a Chinese Restaurant Process or the Forget-Me-Not Process, to pool modes at segmentation time instead of a post-segmentation merge process.

Beyond improving segmentation, there are also possible improvements to learning guarded transitions.  Assuming we had perfect segmentation and mode assignment, we would still not be able to fully capture the guarded transitions of Mario given that our transitions do not have knowledge of Mario's horizontal velocity, nor are they able to learn transitions based on comparisons to arbitrary thresholds.  In some domains, experimentation is possible: we might be able to control the dynamical system in question or to put it into situations where its behavior could be informative.
We would like to explore this to improve the precision of our analysis, either by helping to split truly distinctive merged modes or by testing hypothesized guard conditions. 

\bibliographystyle{named}
\bibliography{ijcai17}

\end{document}